\documentclass{article}
\usepackage{spconf,amsmath,graphicx,hyperref}
\usepackage{cite}
\usepackage{amsmath,amssymb,amsfonts}
\usepackage[table]{xcolor}
\usepackage{algorithmic}
\usepackage{textcomp}
\usepackage{xcolor}
\usepackage{booktabs}
\usepackage{booktabs}
\usepackage{multirow}
\definecolor{oursblue}{RGB}{218,242,252}

\title{RADC: Risk-Aware Dual Caching for Vision-Language Test-Time Adaptation}

\name{
Siyu Huang$^{1,2,*}$ \qquad
Yueyong Chen$^{3,*}$ \qquad
Xuejiao Li$^{2}$ \qquad
Jun Zhou$^{1,\dagger}$%
\thanks{%
\begin{tabular}[t]{@{}l@{}}
$^{*}$ These authors contributed equally to this work.\\
$^{\dagger}$ Corresponding author: Jun Zhou.
\end{tabular}%
}
}

\address{
$^{1}$ Tsinghua Shenzhen International Graduate School,
Tsinghua University, Shenzhen, China \\
$^{2}$ Pengcheng Laboratory, Shenzhen, China \\
$^{3}$ School of Intelligent Systems Engineering,
Sun Yat-sen University, Shenzhen, China
}

\begin{document}
\ninept
\maketitle
\setlength{\abovedisplayskip}{3.5pt plus 1pt minus 1pt}
\setlength{\belowdisplayskip}{3.5pt plus 1pt minus 1pt}
\setlength{\abovedisplayshortskip}{2pt plus 1pt}
\setlength{\belowdisplayshortskip}{2.5pt plus 1pt minus 1pt}
\setlength{\jot}{2pt}
\begin{abstract}
Cache-based test-time adaptation (TTA) for vision-language models is often hindered by background bias in global representations and unreliable entropy-based cache admission under representation variations. To address these limitations, we propose RADC, which enhances prototype learning through reliable dual caching. RADC introduces a Semantic Foreground Cache that aggregates category-consistent spatial evidence from CLIP representations, yielding foreground prototypes that complement the global cache while mitigating background interference. To reliably manage both caches, Gaussian Risk Admission models multi-view representations as diagonal Gaussian distributions and jointly considers class separation and feature uncertainty to prioritize reliable cache candidates. RADC integrates zero-shot logits with complementary global- and foreground-cache predictions for robust inference. Extensive experiments on cross-domain and out-of-distribution benchmarks demonstrate consistent state-of-the-art performance.
\end{abstract}
\vspace{-0.13cm}
\section{Introduction}
\vspace{-0.1cm}
Vision-language models (VLMs), such as CLIP~\cite{radford2021learning}, have demonstrated strong zero-shot recognition capabilities through large-scale image-text pretraining, and can be further adapted to downstream tasks through post-training~\cite{guo2026should,luo2026learning,yan2025deepmoltex,zhang2026kvpo}. Nevertheless, their performance often degrades when the test distribution deviates from the pretraining data. Test-time adaptation 
(TTA)~\cite{zhang2026zo3t} addresses this issue by exploiting unlabeled test samples during inference. 
Existing approaches adapt prompts or prototypes~\cite{shu2022test,zhang2024dual}, 
while recent cache-based methods retrieve historical test representations to 
refine predictions~\cite{karmanov2024efficient,zhang2024boostadapter}. Since cached 
samples are repeatedly involved in subsequent retrieval and prototype construction, 
the quality and reliability of cached representations are critical to effective 
online adaptation.

Despite their effectiveness, existing cache-based TTA methods face two key 
limitations. First, they typically construct caches from global image representations, 
which inevitably entangle category-relevant foreground information with 
sample-specific backgrounds. Consequently, irrelevant contextual patterns may 
accumulate in cached prototypes and repeatedly bias subsequent predictions. 
Second, cache admission is commonly determined by prediction confidence, typically 
favoring low-entropy samples~\cite{zhang2024dualmemory,karmanov2024efficient,
zhang2024dual,zhang2024boostadapter}. 
However, sample-wise entropy may favor overconfident predictions, motivating reliable caching that exploits complementary visual evidence while accounting for representation uncertainty under view and localization variations.

To this end, we propose RADC, a risk-aware 
dual-cache adaptation framework for vision-language models. 
RADC complements the global cache with a Semantic Foreground 
Cache (SFC), forming a dual-cache architecture that preserves both holistic and 
category-relevant visual evidence. Given the candidate classes predicted for each 
test view, SFC first filters out hypotheses that are semantically inconsistent with 
the Top-1 prediction. It then independently localizes each retained class using 
CLIP spatial contributions and aggregates the resulting class-wise masks to extract 
foreground representations. In this way, the foreground cache provides 
category-relevant prototypes complementary to the global cache while reducing 
persistent background interference.

Reliable dual caching further requires identifying which representations are 
sufficiently stable to contribute to prototype construction. We therefore introduce 
Gaussian Risk Admission (GRA), which models multi-view representations as 
diagonal Gaussian distributions and evaluates each cache candidate by jointly 
considering class separation and feature uncertainty. For foreground representations, 
GRA further incorporates localization uncertainty induced by spatial selection. 
The resulting risk scores govern admission and retention in both the global and 
foreground caches, prioritizing reliable representations for subsequent retrieval 
and prototype construction. By combining complementary dual-cache evidence with 
risk-aware admission, RADC enables robust test-time inference under distribution 
shifts.

In summary, our main contributions are as follows:
(1) We propose RADC, a risk-aware dual-cache framework that complements global 
representations with category-relevant foreground evidence for robust vision-language 
test-time adaptation;
(2) We introduce SFC to construct foreground prototypes from semantically consistent 
spatial evidence, together with GRA to reliably manage both caches by jointly 
considering class separation and representation uncertainty;
(3) Extensive experiments on cross-dataset and out-of-distribution benchmarks 
demonstrate consistent state-of-the-art performance with both RN50 and ViT-B/16 
backbones.

\begin{figure*}[t]
    \centering
    \includegraphics[width=\textwidth]{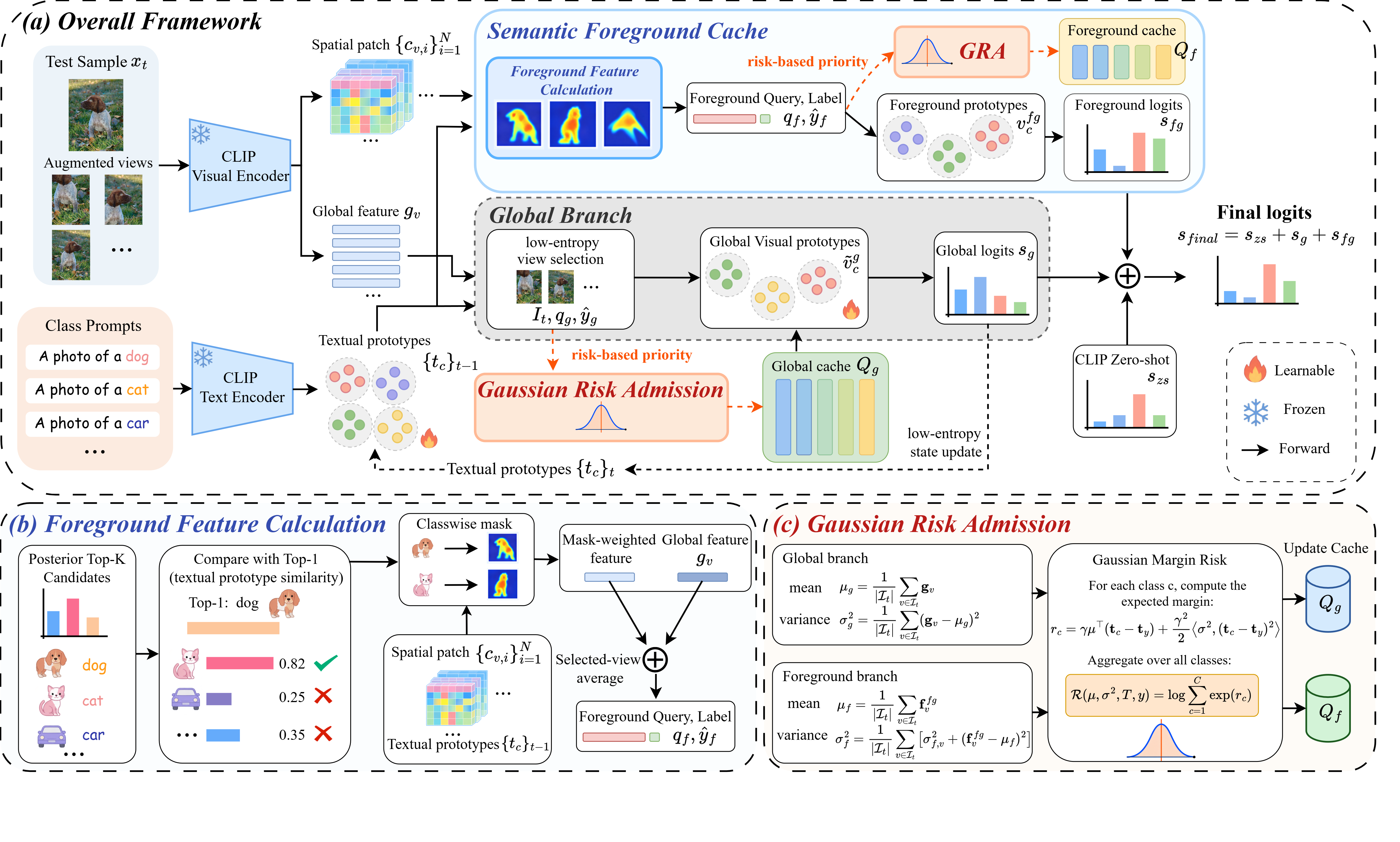}
    \vspace{-1.3cm}
    \caption{
        Overview of RADC.
        (a) RADC complements the global cache with SFC and combines zero-shot, global, and foreground logits
        for prediction.
        (b) SFC localizes category-consistent regions from spatial CLIP
        features to construct foreground queries.
        (c) GRA evaluates class separation and
        feature uncertainty to manage both global and foreground caches.
    }
    \label{fig:framework}
    \vspace{-0.5cm}
\end{figure*}
\vspace{-0.15cm}
\section{Method}
\vspace{-0.1cm}
The overall framework of RADC is illustrated in
Fig.~\ref{fig:framework}. After introducing the necessary preliminaries
in Sec.~\ref{sec:preliminaries}, we present the Semantic Foreground
Cache in Sec.~\ref{sec:sfc} and describe the Gaussian Risk Admission in
Sec.~\ref{sec:gra}.
\vspace{-0.2cm}
\subsection{Preliminaries}
\label{sec:preliminaries}
\noindent\textbf{Zero-shot CLIP.}
CLIP~\cite{radford2021learning} consists of a visual encoder
$E_v(\cdot)$ and a text encoder $E_t(\cdot)$ that align images and
natural-language descriptions in a shared embedding space.
Given a test image $x$ and the prompt $\mathcal{T}_c$ of class $c$,
their normalized representations are
$f=E_v(x)$ and $t_c=E_t(\mathcal{T}_c)$.
For a $C$-class task, the zero-shot logit and prediction probability are:
\begin{equation}
s_c^{\mathrm{zs}}
=
\frac{f^\top t_c}{\tau},
\qquad
p_{\mathrm{zs}}(y=c\mid x)
=
\frac{\exp(s_c^{\mathrm{zs}})}
{\sum_{j=1}^{C}\exp(s_j^{\mathrm{zs}})}.
\label{eq:clip_prediction}
\end{equation}
where $\tau$ denotes the temperature parameter.

\noindent\textbf{Global Prototype Cache.}
DPE~\cite{zhang2024dual} maintains a class-wise priority cache
$\mathcal{Q}_c^g$ that retains low-entropy global features assigned by
pseudo-labels. The corresponding prototype is
\begin{equation}
v_c^g
=
\frac{1}{|\mathcal{Q}_c^g|}
\sum_{f_i\in\mathcal{Q}_c^g}f_i.
\label{eq:global_visual_prototype}
\end{equation}
DPE adapts the textual and visual prototypes through zero-initialized
residuals, $\widetilde t_c=\operatorname{Norm}(t_c+\Delta t_c)$ and
$\widetilde v_c^g=\operatorname{Norm}(v_c^g+\Delta v_c^g)$, optimized
by one-step confidence minimization and cross-modal alignment. For a
global query $q_g$, the cache score is
$s_c^g=\alpha_g\exp[-\beta_g(1-q_g^\top\widetilde v_c^g)]$, where
$\alpha_g$ and $\beta_g$ control its scale and sharpness. We retain this
prototype construction and residual optimization.

\begin{table*}[t]
    \centering
    \caption{
        Top-1 accuracy (\%) on cross-dataset generalization using
        ResNet-50 and ViT-B/16 CLIP visual backbones.
        The best result in each column is highlighted in \textbf{bold},
        while the second-best result is \underline{underlined}.
    }
    \label{tab:cross_dataset_generalization}

    \scriptsize
    \setlength{\tabcolsep}{2.3pt}
    \renewcommand{\arraystretch}{1.08}

    \begin{tabular}{@{}lccccccccccc@{}}
        \toprule
        Method
        & Aircraft
        & Caltech
        & Cars
        & DTD
        & EuroSAT
        & Flowers102
        & Food101
        & Pets
        & SUN397
        & UCF101
        & Average \\
        \midrule

        \textbf{CLIP-RN50}~\cite{radford2021learning}
        & 15.66 & 85.88 & 55.70 & 40.37 & 23.69
        & 61.75 & 73.97 & 83.57 & 58.80 & 58.84 & 55.82 \\

        CoOp~\cite{zhou2022learning}
        & 15.12 & 86.53 & 55.32 & 37.29 & 26.20
        & 61.55 & 75.59 & \underline{87.00}
        & 58.15 & 59.05 & 56.18 \\

        TPT~\cite{shu2022test}
        & 17.58 & 87.02 & 58.46 & 40.84 & 28.33
        & 62.69 & 74.88 & 84.49 & 61.46 & 60.82 & 57.66 \\

        DiffTPT~\cite{feng2023diverse}
        & 17.60 & 86.89 & \underline{60.71} & 40.72 & 41.04
        & 63.53 & \textbf{79.21} & 83.40
        & 62.72 & 62.67 & 59.85 \\

        DMN-ZS~\cite{zhang2024dualmemory}
        & \underline{22.77} & 90.14 & 60.02 & 50.41
        & \textbf{48.72} & 67.93 & 76.70 & 86.78
        & 64.39 & 65.34 & \underline{63.32} \\

        TDA~\cite{karmanov2024efficient}
        & 17.61 & 89.70 & 57.78 & 43.74 & 42.11
        & 68.74 & 77.75 & 86.18 & 62.53 & 64.18 & 61.03 \\

        DPE~\cite{zhang2024dual}
        & 19.80 & 90.83 & 59.26 & 50.18 & 41.67
        & 67.60 & 77.83 & 85.97 & 64.23 & 61.98 & 61.94 \\

        BoostAdapter~\cite{zhang2024boostadapter}
        & 18.93 & 88.48 & 59.67 & 43.85 & 44.40
        & 68.25 & \underline{78.78} & 85.75
        & 62.83 & 64.42 & 61.54 \\

        SCA~\cite{guan2026statistics}
        & 20.82 & \underline{91.03} & 59.73 & \underline{53.31}
        & 43.33 & \textbf{71.09} & 76.51 & 86.15
        & \underline{64.50} & \textbf{66.40} & 63.29 \\

        \rowcolor{oursblue}
        \textbf{RADC (Ours)}
        & \textbf{22.95}
        & \textbf{91.28}
        & \textbf{62.72}
        & \textbf{54.37}
        & \underline{46.32}
        & \underline{69.91}
        & 78.66
        & \textbf{88.01}
        & \textbf{65.88}
        & \underline{65.37}
        & \textbf{64.55} \\

        \midrule

        \textbf{CLIP-ViT-B/16}~\cite{radford2021learning}
        & 23.67 & 93.35 & 65.48 & 44.27 & 42.01
        & 67.44 & 83.65 & 88.25 & 62.59 & 65.13 & 63.58 \\

        CoOp~\cite{zhou2022learning}
        & 18.47 & 93.70 & 64.51 & 41.92 & 46.39
        & 68.71 & 85.30 & 89.14 & 64.15 & 66.55 & 63.88 \\

        TPT~\cite{shu2022test}
        & 24.78 & 94.16 & 66.87 & 47.75 & 42.44
        & 68.98 & 84.67 & 87.79 & 65.50 & 68.04 & 65.10 \\

        DiffTPT~\cite{feng2023diverse}
        & 25.60 & 92.49 & 67.01 & 47.00 & 43.13
        & 70.10 & \textbf{87.23} & 88.22
        & 65.74 & 62.67 & 64.92 \\

        DMN-ZS~\cite{zhang2024dualmemory}
        & \underline{30.03} & \textbf{95.38} & 67.96 & 55.85
        & 59.43 & 74.49 & 85.08 & \underline{92.04}
        & 70.18 & \underline{72.51} & 70.30 \\

        TDA~\cite{karmanov2024efficient}
        & 23.91 & 94.24 & 67.28 & 47.40 & 58.00
        & 71.42 & 86.14 & 88.63 & 67.62 & 70.66 & 67.53 \\

        DPE~\cite{zhang2024dual}
        & 28.95 & 94.81 & 67.31 & 54.20 & 55.79
        & 75.07 & 86.17 & 91.14 & 70.07 & 70.44 & 69.40 \\

        BoostAdapter~\cite{zhang2024boostadapter}
        & 27.45 & 94.77 & \underline{69.30} & 45.69
        & \underline{61.22} & 71.66 & \underline{87.17}
        & 89.51 & 68.09 & 71.93 & 68.68 \\

        SCA~\cite{guan2026statistics}
        & 28.50 & 94.85 & 68.49 & \underline{57.09}
        & 57.16 & \underline{76.09} & 86.09 & 91.44
        & \underline{70.27} & \textbf{73.43}
        & \underline{70.34} \\

        \rowcolor{oursblue}
        \textbf{RADC (Ours)}
        & \textbf{32.28}
        & \underline{95.29}
        & \textbf{70.50}
        & \textbf{58.87}
        & \textbf{63.26}
        & \textbf{78.20}
        & 87.16
        & \textbf{92.53}
        & \textbf{71.02}
        & 72.32
        & \textbf{72.14} \\

        \bottomrule
    \end{tabular}
    \vspace{-0.5cm}
\end{table*}
\vspace{-0.2cm}
\subsection{Semantic Foreground Cache}
\label{sec:sfc}
Global cache representations encode the entire image and may therefore
retain sample-specific backgrounds together with category-relevant
content. Once stored, such irrelevant context can repeatedly participate
in cache retrieval and prototype construction, causing background bias
to propagate across the test stream. To alleviate this issue, we propose
a Semantic Foreground Cache (SFC) that complements the global cache with
category-relevant foreground evidence.

Inspired by the spatial decomposition of CLIP representations
in~\cite{gandelsman2024interpreting}, we obtain patch-level
contributions for ViT from selected late self-attention layers following
their decomposition. For CLIP-RN50, whose final image representation is
aggregated by \texttt{AttentionPool2d}, we further decompose this pooling
operation. The contribution of location $i$ in the $v$-th augmented view is formulated as:
\begin{equation}
c_{v,i}^{\mathrm{RN}}
=
\operatorname{Concat}_{h=1}^{H}
\left(
a_{v,h,i}X_{v,i}W_V^h
\right)W_O.
\label{eq:rn_contribution}
\end{equation}
where $a_{v,h,i}$ denotes the attention weight in head $h$, $X_{v,i}$
is the corresponding spatial feature, and $W_V^h$ and $W_O$ are the
value and output projections. We use $\{c_{v,i}\}_{i=1}^{N}$ to
uniformly denote the spatial contributions from either backbone.

Using the adapted text prototypes $\widetilde{t}_c$, we compute the
posterior probability that the $v$-th view belongs to class $c$ as
$p_{v,c}
=\operatorname{softmax}_c
\left(
\gamma g_v^\top\widetilde{t}_c
\right).$
Directly combining several candidate text prototypes may obscure their
semantic differences before the corresponding visual evidence is
identified. We therefore retain the Top-1 prediction and remove
Top-$K$ candidates that are semantically inconsistent with it:
\begin{equation}
\mathcal K_v
=
\{k_{v,1}\}
\cup
\left\{
k\in\mathcal I_v^K\setminus\{k_{v,1}\}
\,\middle|\,
\cos(\widetilde{t}_k,\widetilde{t}_{k_{v,1}})
\geq\tau_s
\right\}.
\label{eq:semantic_filter}
\end{equation}
where $\mathcal I_v^K$ is the Top-$K$ class set, $k_{v,1}$ is the
class with the highest posterior, and $\tau_s$ denotes the semantic
similarity threshold.

Each retained class is then localized independently, rather than being
merged into a single text direction. The resulting masks are fused only
after their spatial responses have been obtained:
\begin{equation}
\begin{gathered}
M_{v,i,k}
=
\operatorname{sigmoid}
\left\{
\kappa
\left[
\frac{
\cos(c_{v,i},\widetilde{t}_k)-\mu_{v,k}
}{
\max(\sigma_{v,k},\epsilon)
}
-\delta
\right]
\right\},\\
M_{v,i}
=
\sum_{k\in\mathcal K_v}
\bar p_{v,k}M_{v,i,k}.
\end{gathered}
\label{eq:classwise_mask}
\end{equation}
Here, $\mu_{v,k}$ and $\sigma_{v,k}$ are the mean and standard
deviation of the spatial responses for class $k$, and
$\bar p_{v,k}$ denotes the posterior normalized over
$\mathcal K_v$. This procedure preserves the spatial evidence of each
candidate class before mask fusion.

The fused mask $M_{v,i}$ identifies foreground regions associated with
the current category hypotheses. We use it to pool the spatial
contributions and combine the resulting foreground evidence with the
original global feature:
\begin{equation}
f_v^{\mathrm{fg}}
=
\operatorname{Norm}
\left[
(1-\lambda_b)g_v+
\lambda_b\operatorname{Norm}
\left(
\sum_{i=1}^{N}M_{v,i}c_{v,i}
\right)
\right].
\label{eq:foreground_feature}
\end{equation}
The residual combination preserves useful global information when the
foreground localization is incomplete or uncertain.

Let $\mathcal I_t$ denote the views with the lowest prediction entropy.
Their foreground features are averaged and normalized to construct
$q_f=\operatorname{Norm}
\bigl(|\mathcal I_t|^{-1}
\sum_{v\in\mathcal I_t}f_v^{\mathrm{fg}}\bigr)$.
The query is assigned to the class with the highest similarity to the
adapted text prototypes, while GRA(Sec.~\ref{sec:gra}) determines its cache priority and
admission.

For each class $c$, the foreground prototype
$v_c^{\mathrm{fg}}$ is obtained by averaging the previously admitted
queries of that class. The historical cache is queried before the
current foreground query is inserted, preventing the current sample
from contributing to its own prediction. The foreground score is first
computed from the similarity between the current query and the
foreground prototype, and is then combined with the zero-shot and
global cache scores for final prediction:
\begin{equation}
\begin{gathered}
s_c^{\mathrm{fg}}
=
\alpha_f
\exp\left[
-\beta_f
\left(
1-q_f^\top v_c^{\mathrm{fg}}
\right)
\right],\\
s_c^{\mathrm{final}}
=
s_c^{\mathrm{zs}}
+
s_c^{g}
+
s_c^{\mathrm{fg}}.
\end{gathered}
\label{eq:foreground_prediction}
\end{equation}
The foreground score is set to zero when class $c$ has no historical
support.
\begin{table*}[t]
    \centering
    \caption{
        Top-1 accuracy (\%) on robustness to natural distribution shifts
        using ResNet-50 and ViT-B/16 CLIP visual backbones.
        The best result in each column is highlighted in \textbf{bold},
        while the second-best result is \underline{underlined}.
    }
    \label{tab:natural_distribution_shift}

    \scriptsize
    \setlength{\tabcolsep}{5.0pt}
    \renewcommand{\arraystretch}{1.08}

    \begin{tabular}{@{}lccccc|cc@{}}
        \toprule
        Method
        & ImageNet
        & ImageNet-A
        & ImageNet-V2
        & ImageNet-R
        & ImageNet-S
        & Average
        & OOD Average \\
        \midrule

        \textbf{CLIP-RN50}~\cite{radford2021learning}
        & 58.16 & 21.83 & 51.41 & 56.15 & 33.37
        & 44.18 & 40.69 \\

        CoOp~\cite{zhou2022learning}
        & 63.33 & 23.06 & 55.40 & 56.60 & 34.67
        & 46.61 & 42.43 \\

        TPT~\cite{shu2022test}
        & 60.74 & 26.67 & 54.70 & 59.11 & 35.09
        & 47.26 & 43.89 \\

        DiffTPT~\cite{feng2023diverse}
        & 60.80 & 31.06 & 55.80 & 58.80 & 37.10
        & 48.71 & 45.69 \\

        DMN-ZS~\cite{zhang2024dualmemory}
        & \underline{63.87} & 28.57 & 56.12 & 61.44 & 39.84
        & 49.97 & 46.49 \\

        TDA~\cite{karmanov2024efficient}
        & 61.35 & 30.29 & 55.54 & 62.58 & 38.12
        & 49.58 & 46.63 \\

        DPE~\cite{zhang2024dual}
        & 63.41 & 30.15 & \underline{56.72} & \underline{63.72}
        & \underline{40.03} & \underline{50.81} & 47.66 \\

        BoostAdapter~\cite{zhang2024boostadapter}
        & 61.08 & \underline{35.12} & 56.14 & 62.66 & 38.87
        & 50.77 & \underline{48.20} \\

        SCA~\cite{guan2026statistics}
        & 63.11 & 30.64 & 56.51 & 62.85 & 39.33
        & 50.49 & 47.33 \\

        \rowcolor{oursblue}
        \textbf{RADC (Ours)}
        & \textbf{64.31}
        & \textbf{35.16}
        & \textbf{57.33}
        & \textbf{63.87}
        & \textbf{40.59}
        & \textbf{52.25}
        & \textbf{49.24} \\

        \midrule

        \textbf{CLIP-ViT-B/16}~\cite{radford2021learning}
        & 66.73 & 47.87 & 60.86 & 73.98 & 46.09
        & 59.11 & 57.20 \\

        CoOp~\cite{zhou2022learning}
        & 71.51 & 49.71 & 64.20 & 75.21 & 47.99
        & 61.72 & 59.28 \\

        TPT~\cite{shu2022test}
        & 68.98 & 54.77 & 63.45 & 77.06 & 47.94
        & 62.44 & 60.81 \\

        DiffTPT~\cite{feng2023diverse}
        & 70.30 & 55.68 & 65.10 & 75.00 & 46.80
        & 62.58 & 60.65 \\

        DMN-ZS~\cite{zhang2024dualmemory}
        & \underline{72.25} & 58.28 & 65.17 & 78.55
        & \underline{53.20} & 65.49 & 63.80 \\

        TDA~\cite{karmanov2024efficient}
        & 69.51 & 60.11 & 64.67 & 80.24 & 50.54
        & 65.01 & 63.89 \\

        DPE~\cite{zhang2024dual}
        & 71.91 & 59.63 & 65.44 & 80.40 & 52.26
        & 65.93 & 64.43 \\

        BoostAdapter~\cite{zhang2024boostadapter}
        & 69.39 & \underline{64.53} & \underline{65.51}
        & \textbf{80.95} & 51.28
        & \underline{66.33} & \underline{65.57} \\

        SCA~\cite{guan2026statistics}
        & 71.75 & 60.33 & 65.38 & \underline{80.85} & 52.50
        & 66.16 & 64.77 \\

        \rowcolor{oursblue}
        \textbf{RADC (Ours)}
        & \textbf{72.59}
        & \textbf{64.63}
        & \textbf{66.04}
        & 80.75
        & \textbf{53.29}
        & \textbf{67.46}
        & \textbf{66.18} \\

        \bottomrule
    \end{tabular}
    \vspace{-0.5cm}
\end{table*}

\begin{table}[t]
    \centering
    \vspace{-0.2cm}
    \caption{
        Ablation study of SFC and GRA on Flowers102, Pets,
        and Cars.
    }
    \label{tab:component_ablation}

    \scriptsize
    \setlength{\tabcolsep}{2.0pt}
    \renewcommand{\arraystretch}{0.98}

    \begin{tabular}{@{}lcccccc@{}}
        \toprule
        Backbone
        & SFC
        & GRA
        & Flowers102
        & Pets
        & Cars
        & Average \\
        \midrule

        \multirow{4}{*}{RN50}
        & -- & --
        & 67.60
        & 85.97
        & 59.26
        & 70.94 \\

        & $\checkmark$ & --
        & 69.43
        & 87.90
        & 62.34
        & 73.22 \\

        & -- & $\checkmark$
        & 69.75
        & 87.54
        & 61.97
        & 73.09 \\

        & $\checkmark$ & $\checkmark$
        & \textbf{69.91}
        & \textbf{88.01}
        & \textbf{62.72}
        & \textbf{73.55} \\

        \midrule

        \multirow{4}{*}{ViT-B/16}
        & -- & --
        & 75.07
        & 86.17
        & 67.31
        & 76.18 \\

        & $\checkmark$ & --
        & 78.12
        & 92.40
        & 70.41
        & 80.31 \\

        & -- & $\checkmark$
        & 77.71
        & 92.18
        & 69.85
        & 79.91 \\

        & $\checkmark$ & $\checkmark$
        & \textbf{78.20}
        & \textbf{92.53}
        & \textbf{70.50}
        & \textbf{80.41} \\

        \bottomrule
    \end{tabular}
    \vspace{-0.5cm}
\end{table}
\vspace{-0.2cm}
\subsection{Gaussian Risk Admission}
\label{sec:gra}
Prediction entropy reflects output confidence but does not capture the
stability of the underlying representation. We therefore propose
Gaussian Risk Admission (GRA) to evaluate each cache candidate using
both its class margin and feature uncertainty.

We model the visual evidence from selected views as a diagonal Gaussian
distribution
$\mathbf z\sim\mathcal N
(\boldsymbol\mu,\operatorname{diag}(\boldsymbol\sigma^2))$.
Given a fixed pseudo-label $y$, we compare it with each class $c$ in
the text embedding space. Let $\mathbf t_c$ and $\mathbf t_y$ denote
their corresponding text prototypes, and define the discriminative
direction between them as
$\mathbf d_c=\mathbf t_c-\mathbf t_y$.
For each class $c$, we define a class-wise risk $r_c$ that accounts for
both the mean class margin and the feature uncertainty along
$\mathbf d_c$. The overall Gaussian risk $\mathcal R$ is then obtained
by aggregating the risks over all classes:
\begin{equation}
r_c
=
\gamma\boldsymbol\mu^\top\mathbf d_c
+
\tfrac{\gamma^2}{2}
\langle\boldsymbol\sigma^2,\mathbf d_c^2\rangle,
\qquad
\mathcal R
=
\log\sum_{c=1}^{C}e^{r_c}.
\label{eq:gaussian_risk}
\end{equation}
In $r_c$, the first term measures the mean logit difference between
class $c$ and the pseudo-label class, while the second penalizes feature
variation along the corresponding discriminative direction. Therefore,
a larger margin from competing classes and smaller feature variation
lead to a lower $\mathcal R$, indicating that the current representation
is more stable around its pseudo-label.

GRA is applied to both the Global Cache and SFC. For the global branch,
$\boldsymbol\mu_g=|\mathcal I_t|^{-1}\sum_{v\in\mathcal I_t}\mathbf g_v$
and
$\boldsymbol\sigma_g^2=|\mathcal I_t|^{-1}
\sum_{v\in\mathcal I_t}
(\mathbf g_v-\boldsymbol\mu_g)^2$,
capturing variation across the selected global views. For the foreground
branch,
$\boldsymbol\mu_f=|\mathcal I_t|^{-1}
\sum_{v\in\mathcal I_t}\mathbf f_v^{\mathrm{fg}}$,
while the variance further includes localization uncertainty introduced
by the semantic mask:
\begin{equation}
\boldsymbol\sigma_f^2
=
\frac{1}{|\mathcal I_t|}
\sum_{v\in\mathcal I_t}
\left[
\boldsymbol\sigma_{f,v}^2
+
(\mathbf f_v^{\mathrm{fg}}-\boldsymbol\mu_f)^2
\right].
\label{eq:foreground_var}
\end{equation}
Here, $\boldsymbol\sigma_{f,v}^2$ is estimated from the Bernoulli mask
variance $M_{v,i}(1-M_{v,i})$. The resulting risks
$\mathcal R_g$ and $\mathcal R_f$ are used as priorities for the Global
Cache and SFC, respectively. Lower-risk entries receive higher retention
priority, reducing the chance of storing unstable representations based
only on instantaneous prediction entropy.
\vspace{-0.1cm}
\section{Experiments}
\subsection{Experiment Settings}
\noindent\textbf{Datasets.}
Following standard test-time adaptation protocols~\cite{shu2022test},
we evaluate our method on the Cross-Dataset (CD) and
Out-of-Distribution (OOD) benchmarks. The CD benchmark comprises
Aircraft~\cite{maji2013fine}, Caltech101~\cite{fei2004learning},
StanfordCars~\cite{krause20133d}, DTD~\cite{cimpoi2014describing},
EuroSAT~\cite{helber2019eurosat},
Flowers102~\cite{nilsback2008automated},
Food101~\cite{bossard2014food}, OxfordPets~\cite{parkhi2012cats},
SUN397~\cite{xiao2010sun}, and UCF101~\cite{soomro2012dataset}.
The OOD benchmark uses ImageNet~\cite{deng2009imagenet} as the
in-distribution reference and evaluates natural distribution shifts on
ImageNet-A~\cite{hendrycks2021natural},
ImageNet-V2~\cite{recht2019imagenet},
ImageNet-R~\cite{hendrycks2021many}, and
ImageNet-Sketch~\cite{wang2019learning}.

\noindent\textbf{Implementation Details.}
We implement our method with frozen CLIP-RN50 and CLIP-ViT-B/16
backbones. For each image, we retain the 10\% lowest-entropy views among
the original and 63 AugMix views. SFC extracts spatial contributions
from the last three ViT attention layers or the RN50
\texttt{AttentionPool2d} layer. We set
$(\tau_s,\kappa,\delta,\epsilon,\lambda_b)
=(0.7,10,0.5,10^{-6},0.1)$, with a foreground cache capacity of 16,
$\alpha_f=2$, and $\beta_f=\beta_g$. All settings are fixed across
datasets, and experiments are conducted on a single NVIDIA RTX 4090 GPU.
\vspace{-0.2cm}
\subsection{Main Results}
\noindent\textbf{CD Transfer.}
Table~\ref{tab:cross_dataset_generalization} shows that RADC achieves the best average accuracies of 64.55\% with RN50 and 72.14\% with
ViT-B/16, surpassing the corresponding runner-up by 1.23 and 1.80
points. It ranks first on six and seven datasets under the two
backbones, demonstrating strong cross-domain transferability.

\noindent\textbf{OOD Generalization.} As shown in Table~\ref{tab:natural_distribution_shift}, RADC achieves
the best overall/OOD averages of 52.25/49.24\% with RN50 and
67.46/66.18\% with ViT-B/16. It leads on all five datasets with RN50 and four with ViT-B/16, demonstrating consistent robustness to distribution shifts.
\vspace{-0.2cm}
\subsection{Ablation Studies}
\noindent\textbf{Component Ablation.}
As shown in Table~\ref{tab:component_ablation}, both SFC and GRA
consistently improve the base model when applied independently.
Combining them achieves the best results, demonstrating their
effectiveness and complementarity.

\begin{figure}[t]
    \centering
    \makebox[\columnwidth][c]{%
        \includegraphics[
            height=0.35\columnwidth,
            keepaspectratio
        ]{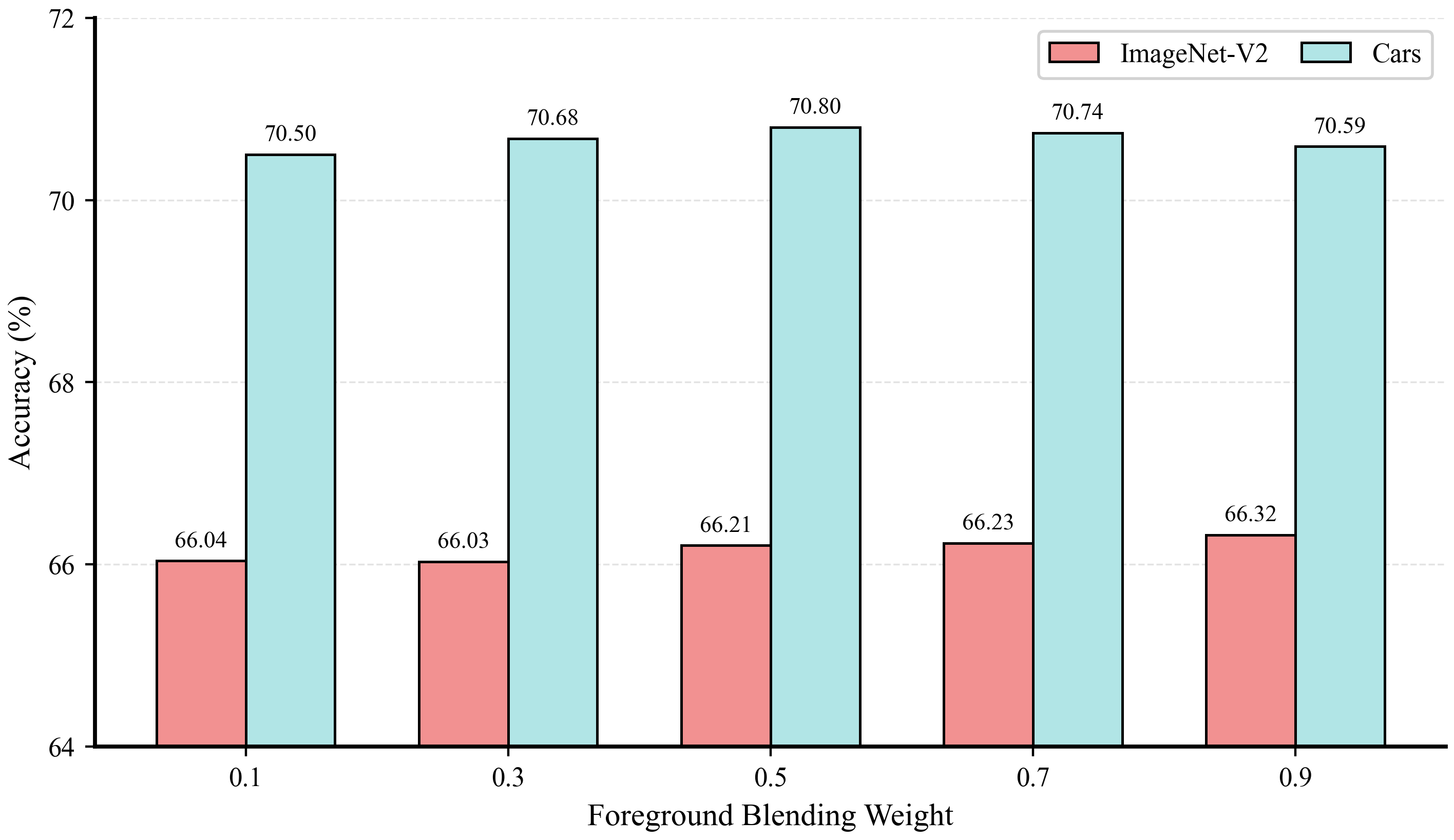}%
        \hspace{1.5mm}%
        \includegraphics[
            height=0.35\columnwidth,
            keepaspectratio
        ]{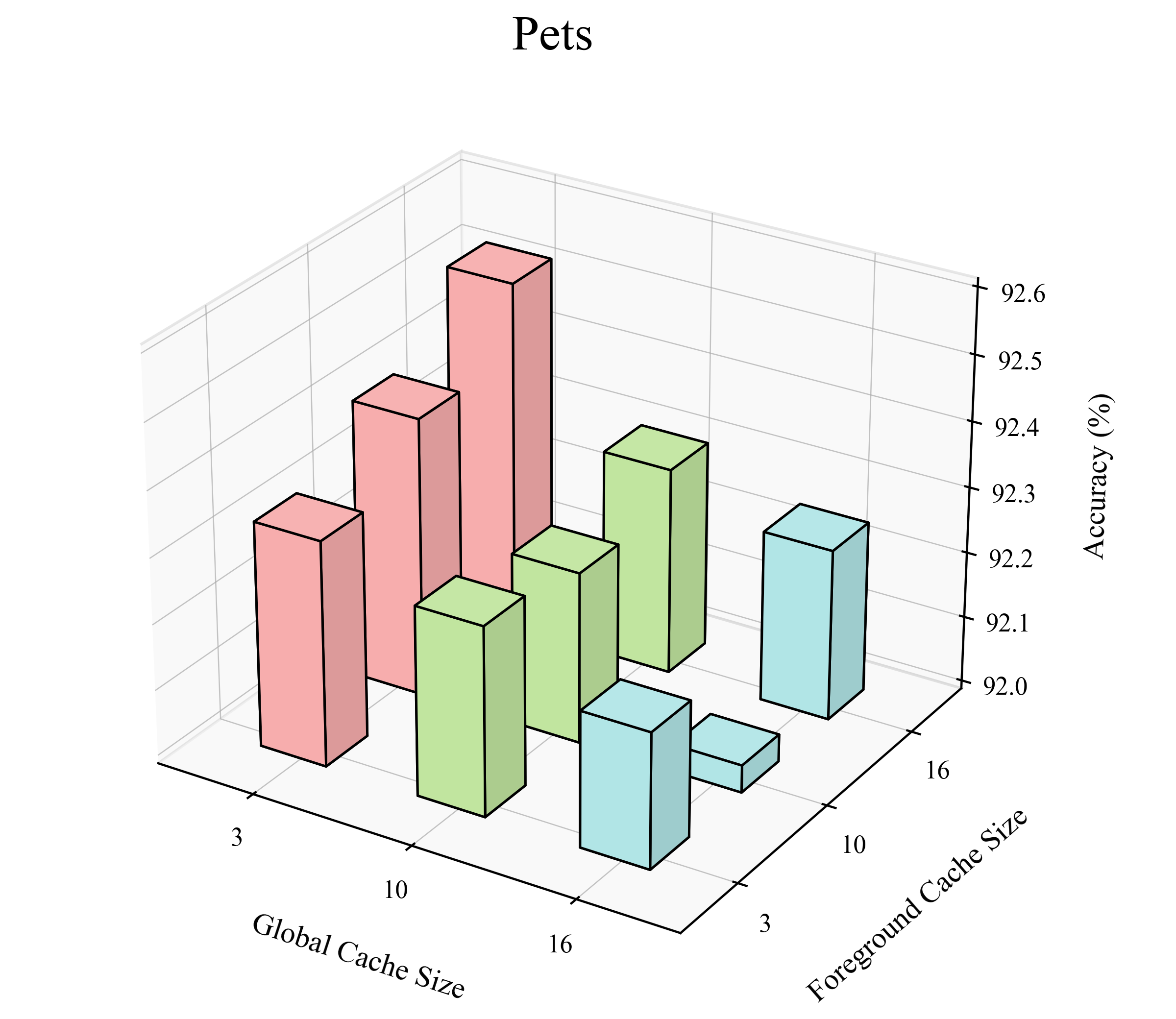}%
    }
    \caption{
        Hyperparameter sensitivity of RADC using ViT-B/16.
        Left: foreground blending weight $\lambda_b$.
        Right: global and foreground cache capacities.
    }
    \label{fig:hyperparameter_sensitivity}
    \vspace{-0.3cm}
\end{figure}

\noindent\textbf{Hyperparameter Sensitivity.}
Figure~\ref{fig:hyperparameter_sensitivity} examines the foreground
blending weight and the capacities of the global and foreground caches.
The performance remains stable across different settings, indicating
that RADC is robust to these hyperparameter choices.
\vspace{-0.25cm}
\subsection{Inference Efficiency Comparison}
Evaluated on ImageNet-A with ViT-B/16 using a single NVIDIA RTX 4090 GPU, our method achieves the highest accuracy while maintaining competitive inference efficiency (Table~\ref{tab:inference_efficiency}).

\begin{table}[t]
    \centering
    \vspace{-0.1cm}
    \caption{
        Comparison with state-of-the-art methods on inference efficiency
        on ImageNet-A.
    }
    \label{tab:inference_efficiency}

    \scriptsize
    \setlength{\tabcolsep}{4.5pt}
    \renewcommand{\arraystretch}{1.08}

    \begin{tabular}{@{}lcccc@{}}
        \toprule
        Method
        & Time
        & Memory
        & Accuracy
        & Gain \\
        \midrule

        CLIP-ViT-B/16~\cite{radford2021learning}
        & 1.13 min
        & 1122 MiB
        & 47.87
        & 0 \\

        \midrule

        DPE~\cite{zhang2024dual}
        & 19.40 min
        & 6466 MiB
        & 59.63
        & +11.76 \\

        SCA~\cite{guan2026statistics}
        & 7.14 min
        & 988 MiB
        & 60.33
        & +12.46 \\

        \rowcolor{oursblue}
        \textbf{RADC (Ours)}
        & 12.26 min
        & 2084 MiB
        & 64.63
        & +16.76 \\

        \bottomrule
    \end{tabular}
    \vspace{-0.5cm}
\end{table}
\vspace{-0.25cm}
\section{Conclusion}
We propose RADC, a risk-aware dual-cache framework for robust vision-language test-time adaptation. By complementing global representations with category-relevant foreground evidence and evaluating cache candidates through Gaussian risk, RADC retains more reliable and complementary representations for prototype learning. This enables robust adaptation while mitigating background interference and representation uncertainty. Extensive experiments on cross-dataset and out-of-distribution benchmarks with RN50 and ViT-B/16 confirm its effectiveness, demonstrating consistent improvements over state-of-the-art methods.

\vspace{3em}
\bibliographystyle{IEEEbib}
\bibliography{strings,refs}

\end{document}